\documentclass[nohyperref]{article}

\usepackage{microtype}
\usepackage{graphicx}
\usepackage{subfigure}
\usepackage{booktabs} 

\usepackage{hyperref}

\newcommand{\ours}{\texttt{FS-CAP}}

\usepackage[accepted]{icml2023}

\usepackage{amsmath}
\usepackage{amssymb}
\usepackage{mathtools}
\usepackage{amsthm}

\usepackage[capitalize,noabbrev]{cleveref}

\theoremstyle{plain}

\theoremstyle{definition}

\theoremstyle{remark}

\usepackage[textsize=tiny]{todonotes}

\icmltitlerunning{Target-Free Compound Activity Prediction via Few-Shot Learning}

\begin{document}

\twocolumn[
\icmltitle{Target-Free Compound Activity Prediction via Few-Shot Learning}



\icmlsetsymbol{equal}{*}

\begin{icmlauthorlist}
\icmlauthor{Peter Eckmann}{cs-ucsd}
\icmlauthor{Jake Anderson}{chemistry-ucsd}
\icmlauthor{Michael K. Gilson}{chemistry-ucsd,skaggs-ucsd}
\icmlauthor{Rose Yu}{cs-ucsd}
\end{icmlauthorlist}

\icmlaffiliation{cs-ucsd}{Department of Computer Science and Engineering, UC San Diego, La Jolla, California, United States}
\icmlaffiliation{chemistry-ucsd}{Department of Chemistry and Biochemistry, UC San Diego, La Jolla, California, United States}
\icmlaffiliation{skaggs-ucsd}{Skaggs School of Pharmacy and Pharmaceutical Sciences, UC San Diego, La Jolla, California, United States}

\icmlcorrespondingauthor{Michael Gilson}{mgilson@health.ucsd.edu}
\icmlcorrespondingauthor{Rose Yu}{roseyu@ucsd.edu}

\icmlkeywords{Machine Learning, ICML}

\vskip 0.3in
]



\printAffiliationsAndNotice{}  

\begin{abstract}
Predicting the activities of compounds against protein-based or phenotypic assays using only a few known compounds and their activities is a common task in target-free drug discovery. Existing few-shot learning approaches are limited to predicting binary labels (active/inactive). However, in real-world drug discovery, degrees of compound activity are highly relevant. We study \textit{Few-Shot Compound Activity Prediction} (\ours{}) and design a novel neural architecture to meta-learn continuous compound activities across large bioactivity datasets. Our model aggregates encodings generated from the known compounds and their activities to capture assay information. We also introduce a separate encoder for the unknown compound. We show that \ours{} surpasses traditional similarity-based techniques as well as other state of the art few-shot learning methods on a variety of target-free drug discovery settings and datasets.
\end{abstract}

\section{Introduction}
A key task in machine learning for drug discovery is to predict the activity of compounds against a target-based or phenotypic assay, reducing the need for expensive lab-based experimental tests \cite{paul2021artificial,vamathevan2019applications}. Most existing methods \cite{ozturk2018deepdta, somnath2021multi, ragoza2017protein, stepniewska2018development, jones2021improved} require information about the target protein, such as amino acid sequence or 3D structure. However, such information is not always available due to experimental difficulties or a lack of mechanistic disease understanding. Indeed, there is increasing interest in target-free drug discovery \cite{haasen2017phenotypic, swinney2020recent} where only a few compounds with weak activity in an experimental assay are known \cite{loew1993strategies, acharya2011recent}. These hit compounds, while not drug candidates themselves, offer a starting point for the discovery of more promising compounds. Traditional methods use chemical similarity, such as the Tanimoto similarity between structural compound fingerprints \cite{bajusz2015tanimoto}, to find new compounds most similar to the hit compounds. However, these compounds are often similarly undesirable as drug candidates, based on the principle that structurally similar compounds have similar properties \cite{johnson1990concepts}.

We cast the problem of target-free compound activity prediction as few-shot learning \cite{wang2020generalizing}, a framework that enables a trained model to generalize to new domains (in this case, assays). Few-shot learning is usually investigated for multi-class classification problems. For drug discovery, these techniques have been applied for binary compound activity prediction \cite{vella2022few, altae2017low}. However, since experimental activity readouts are often continuous \cite{chandrasekaran2021image}, formulation as a binary classification problem requires ad-hoc activity thresholding and is overly simplistic. The few-shot regression problem studied here is more relevant for drug discovery applications \cite{joo2019deep, lenhof2022simultaneous, lee2022metadta}, although it is significantly more challenging \cite{stanley2021fs}.

In this paper, we propose \textit{Few-Shot Compound Activity Prediction} (\ours{}), a model-based few-shot learning approach for target-free compound activity regression. Our model bears some similarity to neural processes (NPs, \citet{garnelo2018conditional}) but with several important differences that are relevant for compound activity prediction. Specifically, we use a deterministic neural encoder to represent context compounds and their activities via a new multiplication-based featurization. We also introduce a separate encoder for the unknown compound to represent its assay-independent binding characteristics. We concatenate these two encodings and feed them to a predictor network to produce a final prediction for the activity of the unknown compound, and train the entire model using mean squared error (MSE).

Despite the rich literature on few-shot classification, few-shot regression remains largely under-explored in drug discovery. 
To the best of our knowledge, only \citet{lee2022metadta} have explored few-shot regression in drug discovery, by applying an Attentive Neural Process (ANP, \citet{kim2019attentive}), a variant of neural processes, to the task. However, many design choices in ANPs are not tailored to compound activity prediction, including their probabilistic framing and lack of unknown compound encoding, which we will show leads to poor performance.
\citet{lee2022metadta} also perform very limited comparison with other few-shot learning methods, and only measure model performance on a single dataset. 

In summary, our contributions include
\begin{itemize}
    \item designing a novel few-shot learning model, \ours{}, that builds on existing neural process designs but with several important architectural and training changes that are specific to drug prediction considerations,
    \item introducing several new datasets and settings to the problem of few-shot compound activity regression that mimic the drug discovery challenges of hit and lead optimization, high-throughput screening, and anti-cancer drug activity prediction, and 
    \item showing that \ours{} outperforms both traditional chemical similarity techniques and modern deep learning-based few-shot learning techniques on this robust set of datasets.
\end{itemize}

\section{Related Work}
We discuss the related work in compound activity prediction and then summarize few-shot learning and its applications to target-free compound activity prediction.

\paragraph{Compound activity prediction.}
Much work focuses on the prediction of compound activities using knowledge of a protein target (e.g. \citet{ozturk2018deepdta, somnath2021multi, ragoza2017protein, stepniewska2018development, jones2021improved}), but such information is not always available in practice \cite{haasen2017phenotypic, swinney2020recent}. In the target-free, or ``ligand-based'' setting, our aim is to use existing compounds (the ``context set'') to predict the activity of unknown compounds (the ``query set'') against new assays. A common computational chemistry technique for this task is to
measure chemical similarity between the context compound(s) and each compound in the query set. This is often performed with binary fingerprints (e.g. \citet{rogers2010extended}), although such structure-based similarity can miss compounds with similar activity but different chemical scaffolds. Therefore, more complex chemical descriptors may also be used, such as polarity, molecular topology, and 3D shape \citep{khan2016descriptors, li2012quantitative, kohlbacher2021qphar, rocs}. 

Machine learning techniques derive molecular representations in a data-driven fashion and thus promise to improve the quality of similarity measurements that use these representations. Much work focuses on the unsupervised learning of molecular representations that can later be used for downstream tasks such as the assessment of compound similarity \citep{jaeger2018mol2vec, huang2021moltrans, li2021molbert, morris2020predicting}. Due to their unsupervised nature, however, similarity measurements between the learned embeddings are not necessarily useful for activity prediction.

\paragraph{Few-Shot Learning.}
Few-shot learning is a framework that enables a trained model to generalize to new domains \cite{wang2020generalizing}. Common techniques include metric-based, optimization-based, and model-based approaches.

Metric-based methods use a learned metric space that is trained specifically to reflect activity differences, as opposed to unsupervised similarity-based methods. \citet{altae2017low} propose an LSTM-based method to iteratively update context compound embeddings, which are used to compute a similarity metric. \citet{schimunekgeneralized} learn a Siamese network-like embedding for compounds in a metric space. The well-known prototypical network \cite{snell2017prototypical} and matching network \cite{vinyals2016matching} techniques have also been proposed for use on molecular graphs \cite{ding2020graph, vella2022few}. However, these techniques only measure similarity between discrete classes (active/active), and cannot use continuous labels. This is problematic when the difference between weakly and highly active compounds is critical, therefore reducing the real-world applicability of such techniques \cite{stanley2021fs, lee2022metadta, lenhof2022simultaneous, joo2019deep}. Indeed, one of the main challenges of drug discovery is to optimize weakly active compounds into highly active ones \cite{hughes2011principles}, yet binary methods like the ones above can make no such distinction.

Optimization-based techniques use gradients computed on the context set to adapt the weights of a ``base'' model, and then apply this adapted model to the query set. Techniques in this area include the LSTM meta-learner \cite{ravi2016optimization}, which uses a separate ``learner'' network to adapt the weights of the main network. \citet{nguyen2020meta} proposed the use of model-agnostic meta-learning (MAML, \citet{finn2017maml}) for few-shot binary compound activity prediction, which finds a set of model parameters that can most quickly be fine-tuned to new tasks.

Instead of updating network weights during test time, model-based approaches take both the query and context set as inputs to a single model. For example, MetaNets \cite{munkhdalai2017metanets} use a memory module coupled with both a base and meta-learner to generate network weights adapted to a new task. Another method, Non-Gaussian Gaussian Processes (NGGPs, \citet{sendera2021nggp}), expands on previous approaches \cite{tossou2019adaptive, rothfuss2021pacoh} that use GPs for few-shot learning by parameterizing the Gaussian posterior with a normalizing flow. However, neither the optimization-based nor the model-based techniques have been applied to few-shot compound activity regression.

Neural processes (NPs, \citet{garnelo2018neuralprocesses}), as well as their variants like attentive neural processes (ANPs, \citet{kim2019attentive}), combine GPs and neural networks for few-shot learning. To the best of our knowledge, the prediction of continuous compound activity values in the few-shot setting has been explored only once in the literature using the ANP-based MetaDTA \cite{lee2022metadta}. However, they include a limited number of experimental settings and baseline comparisons to other few-shot learning models. We propose a novel architecture with some similarity to neural processes but with several important modifications tailored to the prediction of compound activities, and perform a more rigorous comparison across multiple datasets.

\section{Methodology}
\begin{figure*}[t!]
\vskip -0.1in
\begin{center}
\centerline{\includegraphics[width=0.95\textwidth]{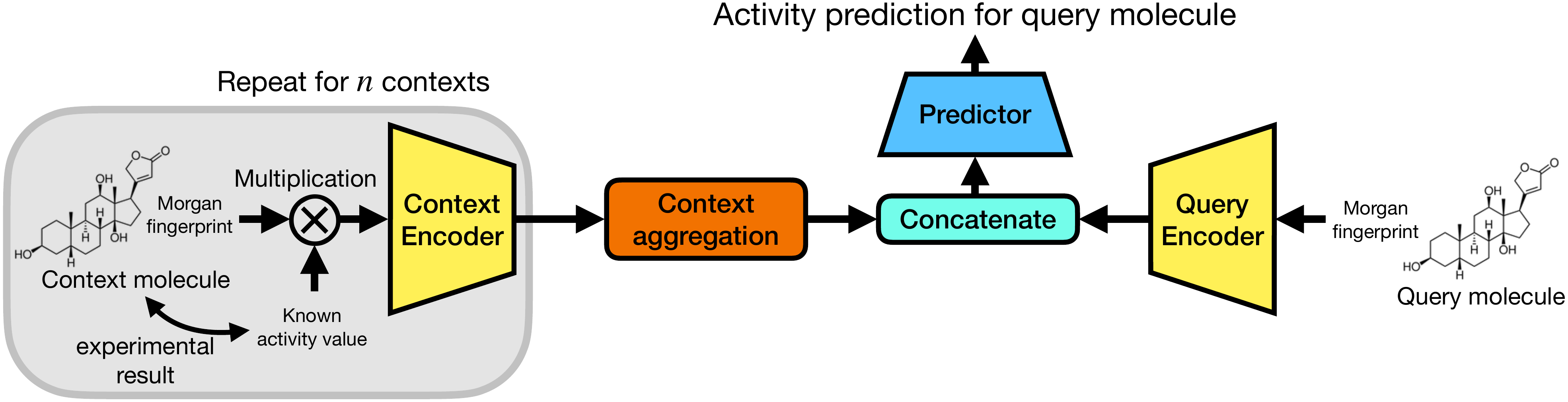}}
\caption{\textbf{Overview of the \ours{} architecture.} The context encoder (left) receives the Morgan fingerprint of each context compound multiplied by its associated activity value. A final context encoding is produced by aggregating the individual encodings of each context compound. The query encoder (right), which has different weights, receives the Morgan fingerprint of the query compound. A predictor network receives the concatenated outputs of each encoder and produces a final scalar activity prediction of the query compound.}
\label{architecture-fig}
\end{center}
\vskip -0.3in
\end{figure*}

We cast the problem of compound activity prediction in new assays given known compounds as a few-shot regression task. To address this problem, we introduce \ours{}, which is summarized in Figure \ref{architecture-fig}.

\paragraph{Problem statement.}
We seek to predict the activity of a ``query'' compound in a new assay, given only a small set of ``context'' compounds and their activities in the same assay. 


Mathematically, suppose our training dataset consists of $K$ different assays. Each assay $k$ consists of $N$ different compounds that are measured against it, $M_k := \{m_1, \cdots, m_{N}\}$. The experimentally measured activity of a molecule $m$ against an assay $k$ is defined as $\pi_k(m) \in \mathbb{R}$. In training, we take a query molecule $m_q$ that is an element of some $M_k$ and aim to predict its activity $\pi_k(m_q)$. To aid in prediction, we randomly sample $n$ context examples from the same assay, $C_k = \{(m_i, \pi_k(m_i))\}_{i=1}^n$, where each $m_i$ is randomly sampled from $M_k$. $n$ must be $\leq N$, and typically it is a small number, hence few-shot. Then, we train the model $f$ to predict the activity of the query molecule given the context set, i.e. $f(m_q, C_k) = \hat{\pi}_k (m_q) \approx \pi_k(m_q)$.

In testing, our model has a similar task, which is to predict the activity of a query molecule given some context set. However, the query and context set come from an assay not seen in training, meaning we measure the ability of the model to adapt its predictions to an unseen assay.

\paragraph{Architecture.}
We employ two separate encoders, a query encoder $f_q$ and a context encoder $f_c$. Consider a single assay $k$. We encode the query molecule $m_q \in M_k$ and elements of the context set $(m_i, \pi_k(m_i)) \in C_k$ as follows:
\begin{equation}
    f_q: m_q \mapsto x_q, \quad f_c: (m_i, \pi_k(m_i))) \mapsto r_i
\end{equation}
where $r_i$ is a representation of the $i$-th context example. The query encoder learns to encode the query molecule into a representation $x_q$, that is useful for predicting its activity. The context encoder learns to capture some information about assay $k$ from each example in the context set. To aggregate each individual context encoding $r_i$ into a single real-valued vector $x_c$ that represents the context set as a whole, we take the average across each $r_i$:
\begin{equation}
    x_c = \frac{1}{n} \sum_{i=1}^n r_{i}.
\end{equation}
This maintains permutation invariance, as desired, since the order of the contexts should not affect their encoding. More complex aggregation techniques, such as self-attention, did not lead to improved performance (Table \ref{variation-table}). 

The predictor network $g$ combines both encodings to generate an activity prediction for the query molecule:
\begin{equation}
g: x_c \oplus x_q \mapsto \hat{\pi}_k(m_q)
\end{equation}
where $\oplus$ denotes vector concatenation.

We represent molecules using their 2048-bit Morgan fingerprints \cite{rogers2010extended}. $f_c$, $f_q$, and $g$ are all multilayer perceptrons with ReLU activations. To pass both the context compound and its measured activity value to $f_c$, we multiply the measured activity scalar with the Morgan fingerprint. Specifically, $f_c$ receives the following vector:
\begin{equation}
    \text{Morgan}(m_i) \cdot \pi_k(m_i)
\end{equation}
Since Morgan fingerprints are substructure-based, i.e. each element in the vector has a 1 bit if there is a certain substructure present and 0 otherwise, and substructures are known to contribute directly to binding characteristics, this featurization may make it easier for the model to learn which substructures contribute how much to activity. We later confirm this intuition by comparing our proposed multiplication approach with the more traditional concatenation of fingerprint and activity values (Table \ref{variation-table}).

\paragraph{Differences to neural processes.}
Although our architecture builds on neural processes (NPs, \citet{garnelo2018neuralprocesses}) and attentive neural processes (ANPs, \citet{kim2019attentive}) such as MetaDTA \cite{lee2022metadta}, it differs in several important aspects that are specific to the few-shot compound activity regression task. First, both NPs and ANPs are based upon a probabilistic framework, which would theoretically allow for the prediction of a distribution of possible activities for a given compound. However, such distributions are not very relevant in drug discovery, where one almost always works with point estimates of compound activity except perhaps in the special case of compound toxicity \cite{lazic2021quantifying}. Avoiding a probabilistic framework stabilizes training, and allows us to simply minimize the mean squared error loss.

Second, NPs and ANPs do not perform query encoding, meaning the query features are fed directly along with the context embedding to the predictor network. However, in drug discovery, there are useful query features that may be extracted entirely independently of any assay, such as compound shape and electrostatics. Allowing the model to encode the query compound in a distinct query encoder, prior to receiving any assay information, is a novel step that appears to improve prediction performance over baselines that use no such encoding (Table \ref{variation-table}).

Third, instead of concatenating the features of the context compound with its activity value, as in NPs or ANPs, we multiply the two before feeding into the context encoder, as described above. This novel featurization, which is only possible due to the unique binary nature of molecular fingerprints and the scalar nature of the activity value, appears to be more effective than concatenation (Table \ref{variation-table}).

\paragraph{Training.}
We use a large assay dataset for training, but set aside some of these assays for testing. We train the model in an end-to-end fashion with Mean Squared Error (MSE), with the loss for each epoch defined as
\begin{equation}
    \mathcal{L} = \frac{1}{K} \sum_{k=1}^K \left( \frac{1}{N} \sum_{i=1}^{N} (\pi_k(m_i) - \hat{\pi}_k(m_i))^2 \right)
\end{equation}
where each $m_i \in M_k$ is a query molecule. 

\section{Experiments}
\subsection{Tasks}
\begin{table*}
\vskip -0.2in
\caption{\textbf{Summary of datasets.} We report the number of assays in each dataset, the number of these assays excluded from training and used for testing, and the number of unique compounds present across all assays in the dataset. We also report the source and access date of the dataset, if applicable.}
\label{dataset-table}
\setlength{\tabcolsep}{3pt}
\vskip 0.15in
\begin{center}
\begin{small}
\begin{sc}
\begin{tabular}{lccccr}
\toprule
& & & Unique & &\\
Dataset & Total assays & Test assays & compounds & Source & Date\\
\midrule
PubChem BioAssays & & & &\\
(PubChemBA) & 98,593 & 1,000 & 1,108,355 & \citet{wang2012pubchem} & 18 Dec. 2022\\
\hline
BindingDB & 4,807 & 100 & 1,013,354 & \citet{gilson2016bindingdb} & 1 Dec. 2022\\
\hline
Cancer Cell Line & & & &\\
Encyclopedia (CCLE) & 275 & 275 & 24 & \citet{barretina2012cancer} & N/A\\
\hline
PubChem High-Throughput\\
Screening (PubChemHTS) & 100 & 100 & 34,716 & \citet{wang2012pubchem} & 23 Dec. 2022\\
\bottomrule
\end{tabular}
\end{sc}
\end{small}
\end{center}
\vskip -0.1in
\end{table*}

We use four different datasets (Table \ref{dataset-table}) to test \ours{} and baseline methods on three tasks related to drug discovery.
\begin{itemize}
    \item \textbf{Hit and lead optimization:} In this scenario, one wishes to use knowledge of a few compounds with modest experimentally determined activities in a binding or phenotypic assay to predict the activities of additional candidate compounds. For this task, we train and test all methods on the BindingDB and PubChem BioAssay (PubChemBA) datasets, which contain continuous activity values across many different assays.
    
    \item \textbf{High-throughput screening:} In high-throughput screening (HTS), large numbers of compounds are assayed to provide a binary active/inactive label. In this scenario, one again has knowledge of a few compounds with modest activities in an assay of interest, but now the goal is to classify a large number of candidate compounds as active or inactive in the assay. Success in this task would provide the ability to use a small amount of data to guide the selection of a compound library for HTS that will have an enhanced fraction of novel actives than a library of randomly chosen compounds. To model this task, we train all methods on the PubChemBA dataset, but treat their outputs as unnormalized probabilities to compute binary classification metrics. We test all methods on the PubChem High-Throughput Screening (PubChemHTS) dataset, which contains binary activity classifications for compounds in PubChem assays marked as ``Screening.'' This dataset contains an entirely separate set of assays from the continuous ones of PubChemBA.
    
    \item \textbf{Anti-cancer drug activity prediction:} We explore whether a model trained on the PubChemBA dataset generalizes to the prediction of compound activity against cancer cell lines. For this task, we use the Cancer Cell Line Encyclopedia (CCLE), which contains IC50 measurements for 24 drugs against 275 patient-derived cancer cell lines. We further probe the biological understanding of the trained models with additional challenges on this dataset involving the generated context encodings.
\end{itemize}

We defer further dataset and preprocessing details to Appendix \ref{appendix-dataset}. We also include additional experimental results on the FS-Mol dataset \cite{stanley2021fs} in Appendix \ref{appendix-results}. 
For all datasets, assay data were expressed as $\log_{10}$ of the  activity in nanomolar (nM) units.

\subsection{Baselines}
As baselines for comparison, we include Tanimoto fingerprint similarity (a widely used traditional technique from computational chemistry) and several state-of-the-art approaches in few-shot learning. We applied both optimization-based (MAML, \citet{finn2017maml}) and model-based (MetaNet, \citet{munkhdalai2017metanets}; ANP, \citet{kim2019attentive}) methods to the regression of compound activities. We omit similarity-based methods (e.g. \citet{snell2017prototypical, vinyals2016matching}) as they require binarizing the activity data of the context compounds, making for an unfair comparison. Details on the training and implementation of \ours{} and baselines are reported in Appendix \ref{appendix-details}.
\begin{itemize}
    \item \textbf{Tanimoto similarity.} Traditional molecular structure-based similarity measure based on binary Morgan fingerprints \citep{rogers2010extended, bajusz2015tanimoto}. When given multiple context compounds, we use the highest similarity score between each of the contexts and the query.
    \item \textbf{MolBERT + attentive neural process (ANP).} Combines MolBERT, which is a start-of-the-art sequence-based molecular featurizer for property prediction tasks \cite{li2021molbert}, with an attentive neural process model \cite{kim2019attentive} for the few-shot prediction of activity values.
    \item \textbf{Non-Gaussian Gaussian process (NGGP)} \cite{sendera2021nggp}. Expands on basic Gaussian process techniques for few-shot learning by modeling the posterior distribution with an ODE-based normalizing flow.
    \item \textbf{MetaNet} \cite{munkhdalai2017metanets}. Uses two separate learners, the base learner and the meta-learner which utilizes a memory mechanism, to quickly adapt to new tasks in the few-shot setting via fast parameterization.
    \item \textbf{Model-agnostic meta-learning (MAML)} \cite{finn2017maml}. Learns a model that can quickly adapt to a new task by training on a small set of context examples. For this paper, we use a simple multilayer perceptron that takes a Morgan fingerprint as input for the base model.
    \item \textbf{MetaDTA} \cite{lee2022metadta}. Applies attentive neural processes to the few-shot regression of continuous activity values. We use the MetaDTA(I) variant because its performance is superior to that of the other reported variants.
\end{itemize}

\subsection{Hit and lead optimization}
\label{sec-scoring}
\begin{table*}
\vskip -0.2in
\caption{\textbf{Average per-assay correlation.} Mean Pearson's $r$ between predicted and ground-truth compound activity values across all test-set assays in PubChemBA and BindingDB. To mimic a hit/lead optimization task, where compounds with high activity are not known, we only sampled context compounds with $>10~\mu M$ activity values. For each method, a separate model was trained on each dataset and for each different number of context compounds. We report the mean $\pm$ one standard deviation from three independent training runs with random seeds for the top three baselines. Due to computational constraints, we report results for all other baselines from one training run.}
\label{corr-table}
\vskip 0.15in
\setlength{\tabcolsep}{4pt}
\begin{center}
\begin{small}
\begin{sc}
\begin{tabular}{l|cccc|cccc}
\toprule
Dataset & \multicolumn{4}{c}{PubChemBA} & \multicolumn{4}
{c}{BindingDB}\\
\# context compounds & 1 & 2 & 4 & 8 & 1 & 2 & 4 & 8\\
\midrule
Tanimoto similarity & 0.01 & 0.13 & 0.21 & 0.27 & -0.08 & 0.05 & 0.13 & 0.18\\
MolBERT + ANP & -0.01 & 0.23 & 0.22 & 0.17 & 0.09 & 0.10 & 0.10 & 0.12\\
NGGP & 0.17 & 0.20 & 0.25 & 0.30 & 0.12 & 0.17 & 0.17 & 0.18\\
MetaNet & 0.02 & 0.05 & 0.06 & 0.02 & 0.06 & -0.01 & 0.05 & 0.09\\
MAML & $0.39_{\pm 0.00}$ & $0.39_{\pm 0.01}$ & $0.39_{\pm 0.00}$ & $0.41_{\pm 0.00}$ & $0.34_{\pm 0.02}$ & $0.35_{\pm 0.02}$ & $0.36_{\pm 0.00}$ & $0.36_{\pm 0.00}$\\
MetaDTA & $0.44_{\pm 0.00}$ & $0.45_{\pm 0.00}$ & $0.45_{\pm 0.00}$ & $0.45_{\pm 0.01}$ & $0.36_{\pm 0.01}$ & $0.36_{\pm 0.00}$ & $0.35_{\pm 0.00}$ & $0.34_{\pm 0.01}$\\
\midrule
FS-CAP & $\mathbf{0.48_{\pm 0.01}}$ & $\mathbf{0.48_{\pm 0.01}}$ & $\mathbf{0.49_{\pm 0.00}}$ & $\mathbf{0.49_{\pm 0.01}}$ & $\mathbf{0.38_{\pm 0.01}}$ & $\mathbf{0.38_{\pm 0.00}}$ & $\mathbf{0.38_{\pm 0.02}}$ & $\mathbf{0.39_{\pm 0.01}}$\\
\bottomrule
\end{tabular}
\end{sc}
\end{small}
\end{center}
\vskip -0.1in
\end{table*}

To explore the applicability of few-shot learning methods to the hit and lead optimization settings, we compare \ours{} with baseline methods on the few-shot prediction of compound activity values against assays in PubChemBA and BindingDB. Compounds with high activity are often not known at the hit stage, so we only sampled context compounds (in both training and testing) that have activity values (i.e. effective concentrations) $>10~\mu M$, which is typical of hit compounds \cite{zhu2013hit}. Note that a higher effective concentration means lower activity. Following training, we test each method against the assays in the held-out test set. Thus, each test set compound was treated as a query compound, with each query being used with a context set of 1-8 compounds randomly sampled from all compounds against the same assay as the query.

Table \ref{corr-table} reports the mean correlation of the predicted and ground truth activity values across all test-set assays for each method. Pearson's correlation coefficient measures the ability of each method to differentiate between compound activities against the same assay and is a standard metric in the literature \cite{jones2021improved, wang2021resatom}. Other metrics, such as MSE, may appear favorable even if a method makes the same prediction for all compounds against a given assay. 

As shown, \ours{} consistently outperforms Tanimoto similarity, the de facto standard in medicinal chemistry, as well as deep learning-based few-shot learning baselines, across datasets and for different numbers of context compounds. This suggests that \ours{} may be useful for hit and lead optimization, as it is the most successful in predicting the activities of unknown compounds using only weakly active context compounds. Similar results were obtained when we performed the same study without any limit on the context compound activities, except that the correlation coefficients were higher by about 0.10 (Appendix \ref{appendix-results}). We used the version of the PubChemBA model trained without any context activity limits for Sections \ref{sec-hts} and \ref{sec-cancer}.

\subsection{High-throughput screening}
\label{sec-hts}
\begin{table}
\vskip -0.1in
\caption{\textbf{Average ROC-AUC and enrichment statistics across all high-throughput screening assays.} ROC-AUC measures the ability of each method to classify compounds as active or inactive. Each percentage value indicates the $k$\% enrichment. 8 context compounds were used. We report the mean ($\pm$ one standard deviation for ROC-AUC) from three independent training runs on the PubChemBA dataset with random seeds for the top three baselines.}
\label{hts-table}
\setlength{\tabcolsep}{3pt}
\begin{center}
\begin{small}
\begin{sc}
\begin{tabular}{l|c|ccc}
\toprule
Method & ROC-AUC & 0.5\% & 1\% & 2\%\\
\midrule
Tanimoto similarity & 0.51 & 76\% & 98\% & 120\%\\
MolBERT + ANP & 0.51 & 160\% & 130\% & 130\%\\
NGGP & 0.49 & 150\% & 110\% & 95\%\\
MetaNet & 0.49 & 100\% & 110\% & 120\%\\
MAML & $0.51_{\pm 0.01}$ & \textbf{210\%} & 150\% & 150\%\\
MetaDTA & $0.55_{\pm 0.00}$ & 160\% & 150\% & 140\%\\
\midrule
FS-CAP & $\mathbf{0.57_{\pm 0.00}}$ & 200\% & \textbf{190\%} & \textbf{180\%}\\
\bottomrule
\end{tabular}
\end{sc}
\end{small}
\end{center}
\vskip -0.1in
\end{table}

We evaluated the performance of \ours{} and baseline methods on the few-shot prediction of compound activities in high-throughput screening (HTS) assays from PubChemHTS (Table \ref{hts-table}). While the activity data for a given HTS assay are binary compound labels, more detailed confirmatory (dose-response) studies are often available for selected hit compounds, which can provide context compounds with continuous activity values. In this task, we obtained context compounds via separate dose-response assays not in PubChemHTS, but with the same targets as PubChemHTS assays (see Appendix \ref{appendix-dataset} for details).

For this task, we train all models on PubChemBA. While the models predict a continuous activity value for each query compound, we treat their outputs as unnormalized probabilities (that were inverted, because a low effective concentration corresponds to a high activity), so that classification metrics may be computed from the model output. In other words, we assumed that a high continuous compound activity prediction from the models corresponded to an ``Active'' classification, and vice-versa. Specifically, we measured performance through ROC-AUC using the ground-truth binary activity labels, a standard metric in the HTS literature \cite{triballeau2005virtual}. We also measured performance with $k$\% enrichment, which is the percent increase of actives over the base rate in the top $k$\% of scored compounds, also a standard metric in the HTS literature \cite{lopes2017power}.

We find that \ours{} outperforms baselines both in ROC-AUC and in most enrichment measurements (Table~\ref{hts-table}). This suggests that \ours{} is more capable of predicting compound activities in screening libraries than baseline methods, and maybe the most effective at raising the hit rate of a library selected from a much larger set of compounds to perform more targeted and cost-effective testing.

\subsection{Anti-cancer drug activity prediction}
\label{sec-cancer}
\begin{table*}
\vskip -0.2in
\caption{\textbf{Average correlation per cell line.} Mean Pearson's $r$ between ground truth and predicted drug activity values across all cell lines in the CCLE. Experiments were performed using 1, 2, 4, and 8 context compounds for each method tested. We report the mean $\pm$ one standard deviation from three independent training runs on the PubChemBA dataset with random seeds for the top three baselines.}
\label{cell-line-table}
\begin{center}
\begin{small}
\begin{sc}
\begin{tabular}{l|cccc}
\toprule
\# context compounds & 1 & 2 & 4 & 8\\
\midrule
Tanimoto similarity & 0.17 & 0.28 & 0.33 & 0.36\\
MolBERT + ANP & 0.04 & 0.11 & -0.13 & 0.07\\
NGGP & 0.12 & 0.18 & 0.25 & 0.32\\
MetaNet & -0.25 & 0.39 & 0.22 & -0.04\\
MAML & $0.50_{\pm 0.05}$ & $0.45_{\pm 0.04}$ & $0.47_{\pm 0.04}$ & $0.17_{\pm 0.03}$\\
MetaDTA & $0.52_{\pm 0.03}$ & $0.49_{\pm 0.03}$ & $\mathbf{0.51_{\pm 0.02}}$ & $0.39_{\pm 0.03}$\\
\midrule
FS-CAP & $\mathbf{0.58_{\pm 0.02}}$ & $\mathbf{0.56_{\pm 0.03}}$ & $\mathbf{0.51_{\pm 0.03}}$ & $\mathbf{0.46_{\pm0.03}}$\\
\bottomrule
\end{tabular}
\end{sc}
\end{small}
\end{center}
\vskip -0.1in
\end{table*}

In this task, we train all models on PubChemBA and test them on the prediction of anti-cancer drug activities against patient-derived cancer cell lines in the Cancer Cell Line Encyclopedia (CCLE, \citet{barretina2012cancer}). Context compounds were randomly sampled from all compounds with activity data against a given cell line, and were used to predict the activities against the same cell line of query compounds not in the context set. We report the mean correlation between predicted and experimentally determined IC50 values for drugs across all cell lines.

As shown in Table \ref{cell-line-table}, \ours{} is better than the baseline methods at predicting the phenotypic activities of anti-cancer drugs. Although the number of compounds tested in the CCLE is relatively small, the success of \ours{} in predicting activity values in this dataset, despite being trained only on PubChemBA, suggests that it may learn fundamental relationships between compounds and assays that generalize across datasets.

\begin{table}
\vskip -0.1in
\caption{\textbf{Accuracy of cell line identification using context encodings.} Accuracy scores of logistic regression models trained to classify the cell line based on context encodings generated by each method pretrained on PubChemBA. We included 20 randomly chosen cell lines, and performed 15 trials for each cell line and a number of context compounds, where a trial consisted of encoding randomly sampled context compounds and their associated activities. We trained a separate logistic regression classifier for each method and number of context compounds using 80\% of the available encodings, and computed the reported accuracy scores on the remaining 20\%. A random classifier would have 5\% accuracy.}
\label{classification-table}
\vskip 0.15in
\begin{center}
\begin{small}
\begin{sc}
\begin{tabular}{l|cccc}
\toprule
\# context compounds & 1 & 2 & 4 & 8\\
\midrule
MetaDTA & 5\% & 8\% & 10\% & 27 \%\\
FS-CAP & \textbf{24\%} & \textbf{39\%} & \textbf{56\%} & \textbf{81\%}\\
\bottomrule
\end{tabular}
\end{sc}
\end{small}
\end{center}
\vskip -0.1in
\end{table}

\begin{figure}[ht]
\vskip 0.2in
\begin{center}
\centerline{\includegraphics[width=\columnwidth]{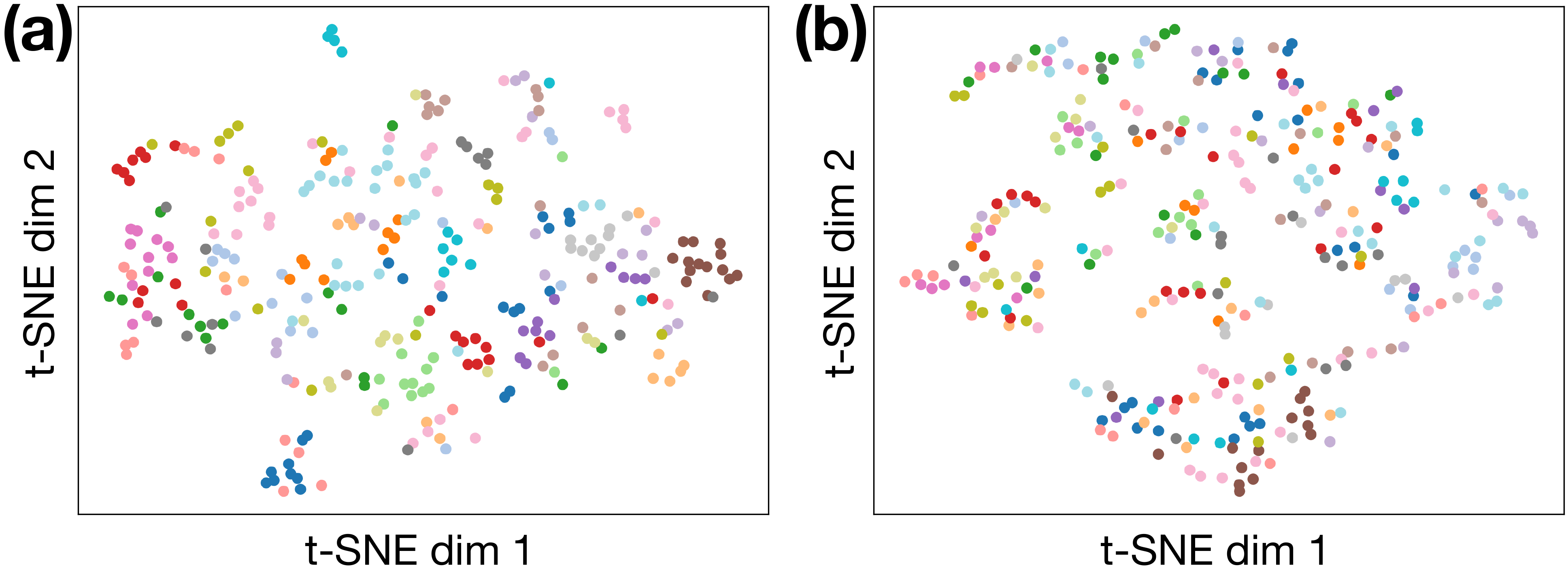}}
\caption{\textbf{t-SNE visualization of context encodings, colored by cell line, generated by (a) FS-CAP and (b) MetaDTA.} Each dot represents one context encoding using 8 randomly sampled context compounds and their associated activities against a given cell line. The color of the dot represents the cell line.}
\label{clustering-fig}
\end{center}
\vskip -0.2in
\end{figure}

We further explored the properties of \ours{}'s trained context encoder using a new classification task. Here, the input was a set of context compounds and their activities against a given cell line, and the output was a prediction of which cell line these activities correspond to. For this task, we applied a simple logistic regression classifier on top of the context encoding generated by \ours{} (i.e. $x_c$). For comparison, we apply a similar approach to the latent path prior of MetaDTA ($z$ in \citet{lee2022metadta}), our most competitive baseline (Table \ref{cell-line-table}).

We randomly selected 20 cell lines in the CCLE. For each of the 20 cell lines and for each number of context compounds, we conducted 15 trials, where each trial consisted of randomly sampling context compounds and their activities against the cell line. As not all compounds have measured activities against all cell lines in the CCLE, we only sampled contexts from the 15 compounds that have experimental activities measured against all 20 cell lines. This prevents the logistic regression classifier from simply learning which compounds were tested against which cell lines. For training the classifier, we used a random 80/20 train/test split, where 80\% of the context encodings and their associated cell lines were used to train the model and the remaining 20\% were used to judge its accuracy.

As shown in Table \ref{classification-table}, the classifier trained on top of \ours{} significantly outperforms that of MetaDTA on the test set, suggesting that the context encodings generated by \ours{} are more meaningful. In addition, Figure \ref{clustering-fig} shows the t-SNE \cite{van2008visualizing} projections of the context encodings generated by \ours{} (left panel) and MetaDTA (right panel) using 8 context compounds. The encodings of \ours{} appear to cluster by cell line (indicated by colors), while the corresponding projections of the MetaDTA encodings appear more scattered, helping to explain the high accuracy of the linear regression classifier trained on \ours{}. Such clustering signifies that \ours{} is able to produce similar encodings of context compounds when their associated activities are derived from the same assay, even if the identity of the context compounds themselves vary. 

Particularly interesting is that such clustering is observed on the cell line dataset despite having been trained on the nonoverlapping PubChemBA dataset. This suggests that training on large assay datasets allows for the extraction of biologically relevant information on how functional drug responses relate to the unique aspects of various cancer cell lines, e.g. type of cancer or mutations present. Along with Table \ref{classification-table}, these results help explain the observed superior performance of \ours{} for compound activity prediction, as a meaningful encoding of assay information is a necessary first step towards predicting the activity of unknown compounds against that assay.

\subsection{Model ablations}
\begin{table}
\vskip -0.1in
\caption{\textbf{Model ablations}. We measure the mean correlation between ground-truth and predicted activities across all test assays in PubChemBA and BindingDB using 8 context compounds. We report the mean $\pm$ one standard deviation from three independent training runs with random seeds.}
\label{variation-table}
\setlength{\tabcolsep}{3pt}
\vskip 0.15in
\begin{center}
\begin{small}
\begin{sc}
\begin{tabular}{l|cc}
\toprule
Ablation & PubChemBA & BindingDB\\\midrule
\textbf{Base model (FS-CAP)} & $\mathbf{0.54_{\pm 0.01}}$ & $\mathbf{0.48_{\pm 0.00}}$\\
No query encoding & $0.53_{\pm 0.00}$ & $0.46_{\pm 0.01}$\\
Concatenated context & $0.53_{\pm 0.00}$ & $0.45_{\pm 0.00}$\\
No context & $0.48_{\pm 0.00}$ & $0.40_{\pm 0.00}$\\
Attentive aggregation & $0.51_{\pm 0.01}$ & $0.30_{\pm 0.01}$\\
\bottomrule
\end{tabular}
\end{sc}
\end{small}
\end{center}
\vskip -0.1in
\end{table}

We report performance metrics of model ablations to the \ours{} architecture in Table \ref{variation-table}. For each ablation, we trained the model and then measured the mean correlation of the predicted and ground truth activity values across all test-set assays in PubChemBA and BindingDB. This experiment is similar to that presented in Section \ref{sec-scoring}, except context compounds are selected at random and not constrained by their activity. 8 context compounds were used for all tests. 

We test the significance of using a separate query encoder network (``Base model''), or feeding the query features directly to the predictor network (``No query encoding''), similar to a typical neural process model. The greater performance of the variation with the query encoder suggests that encoding the query independent of assay information is beneficial for prediction. 

``Concatenated context'' means that we feed the context encoder a binary compound fingerprint concatenated with its associated activity value, instead of multiplying the two. This is similar to a neural process model. This variation shows inferior performance, suggesting that combining the context compound fingerprint and activity value scalar via multiplication is a useful featurization for the activity prediction task. ``No context'' denotes that no context was fed to the model at all, and it made activity predictions based solely on the query compound. ``Attentive aggregation'' means that we applied 4-layer self-attention on the individual context encodings before taking the mean.

\section{Discussion and Conclusions}
The proposed few-shot learning model \ours{} surpasses both a standard chemical similarity metric and prior few-shot learning baselines in multiple tasks of interest in early stage drug discovery. These tasks include prediction of compound activities based on a set of weak-binding context compounds, prediction of screening library compounds as active or inactive, and prediction of antitumor activity in cell-based assays, all performed with models trained on large activity datasets. Together, these results suggest that \ours{} may be broadly useful for target-free, or ligand-based, drug discovery, which has become more common in recent years in comparison to target-based drug discovery that uses protein information \cite{haasen2017phenotypic, swinney2020recent}.

\ours{} may already be useful in its present form as a tool to leverage the limited compound activity data that is typically available in the earliest stages of drug discovery, focusing attention on candidate compounds that are much more likely than randomly chosen compounds to be active in an assay of interest. It thus offers a novel approach to speed drug discovery and reduce its costs. Exploring the use of \ours{} for other compound properties might open further applications. For example, it may find applications in predicting pharmacokinetic parameters of candidate compounds, such as bioavailability and half-life; metabolic susceptibility; and toxicity.

Limitations of the present implementation of \ours{} include its use of a relatively simple molecular representation (Morgan fingerprints), and a context aggregation technique with limited expressiveness. Additionally, the inherent limitations of training on experimental assay data, such as the limited tested dose range \cite{stanley2021fs} or systematic biases in which compounds are tested against which targets, may limit the applicability of few-shot methods like \ours{} trained on these datasets to real-world drug discovery projects.

Future developments could include the exploration of more complex molecular representations (e.g. sequence or graph-based) and the application of more complex context aggregation methods beyond the mean. Finally, research into incorporating target information, when available, with few-shot methods may allow for increased prediction accuracy beyond using target information or context compounds alone.

\section{Acknowledgements}
This work was supported in part by U.S. Department Of Energy, Office of Science, U. S. Army Research Office under Grant W911NF-20-1-0334, and NSF Grants \#2134274 and \#2146343. RY has an equity interest in and is a scientific advisor of Salient Predictions. MKG acknowledges funding from National Institute of General Medical Sciences (GM061300). These ﬁndings are solely of the authors and do not necessarily represent the views of the NIH. MKG has an equity interest in and is a cofounder and scientiﬁc advisor of VeraChem LLC.


\bibliography{main.bib}
\bibliographystyle{icml2023}

\newpage
\appendix
\onecolumn
\section{Dataset details}
\label{appendix-dataset}
\subsection{PubChemBA and BindingDB}
We trained on PubChemBA \cite{wang2012pubchem} and BindingDB \cite{gilson2016bindingdb} for their size, high quality, and broad coverage across many targets and assays. For both datasets, we excluded very small or very large molecules, defined as less than 10 atoms or more than 70. From BindingDB, we recorded activities in nanomolar units from either the $K_D$, $K_i$, IC50, or EC50 columns, if available. Similarly, we used the PubChem ``activity value'', which can be any dose-response activity value (either target-based or phenotypic), normalized to nanomolar units. We used such a broad range of different activity types because all values are similarly determined by an underlying binding mechanism, it increased the amount of data we can train on, and allowed the trained models to generalize to both target-based and phenotypic data types. If no continuous activity value was available for a given molecule, we discarded it. When activity was expressed as an upper or lower bound, we took the bound itself as the known activity. To reduce outlier activity values, we also clipped activity values with log10 nM values of $<-2.5$ or $>6.5$, as values surpassing those limits were rare. Then, we excluded all assays that include less than 10 measured compounds. Assays were defined via protein sequence in BindingDB (although some protein targets may contain data aggregated from multiple experimental assays), and by bioassay (i.e. AssayID) in PubChemBA. We transformed all activity values using the base-10 logarithm, as activity often spans several orders of magnitude.

BindingDB data was taken directly from the file \texttt{BindingDB\_All.tsv} (\url{https://www.bindingdb.org/rwd/bind/chemsearch/marvin/SDFdownload.jsp?download_file=/bind/downloads/BindingDB_All_2D_2023m0.sdf.zip}). PubChemBA data was downloaded via the FTP interface (\url{https://ftp.ncbi.nlm.nih.gov/pubchem/Bioassay/Concise/JSON/}). For each row in the downloaded files, the activity value was taken from the \texttt{PubChem Standard Value} column, and the SubstanceIDs were converted into corresponding SMILES strings via the files available at \url{https://ftp.ncbi.nlm.nih.gov/pubchem/Substance/CURRENT-Full/SDF/}.

\subsection{Cancer Cell Line Encyclopedia}
The Cancer Cell Line Encyclopedia \citep{barretina2012cancer} consists of interaction data of 24 drugs against a wide array of 479 patient-derived cancer cell lines. For this paper, we used the dataset reported in Table S11 of \citet{barretina2012cancer}, and extracted IC50 measurements for each drug measured against each cell line. We excluded compounds with less than 10 or more than 70 atoms, and cell lines with less than 10 drugs with measured activity. We also excluded all compound-activity pairs if there was no continuous activity value reported.

\subsection{PubChemHTS}
Starting with a list of Assay IDs (AIDs) obtained from the search function at \url{https://pubchem.ncbi.nlm.nih.gov/}, we downloaded the top 100 AIDs with the highest number of tested substances with ``BioAssay Type'' equal to ``Screening'' and a linked ``Protein Target'' section in the ``BioAssay Record.'' For each linked protein in a given Screening assay, we obtained continuous activity values to be used as context compounds via the protein's ``Chemicals and Bioactivities'' section in PubChem. As we used these compounds and their activities for context compounds in our experiments, we excluded all proteins, and therefore assays, with less than 10 tested compounds with continuous activity values. After obtaining the context compounds, we downloaded the datatable for each assay, which contained Compound IDs (which were linked to SMILES strings, to be used as query compounds in our tests, via the PubChem API) and binary compound activity classifications (``Active'' or ``Inactive'' in the datatable file, to be used for computing ROC-AUC and enrichment scores).

\section{Implementation details}
\label{appendix-details}

Unless specifically stated, all baselines were trained with the same molecular representation (2048-bit Morgan fingerprints with a radius of 3). For \ours{} and all baseline methods, we tuned hyperparameters once for each model in the PubChemBA task discussed in Section \ref{sec-scoring} using 8 context compounds, except without limiting the context activity range, and used the same hyperparameters for all other datasets and subsequent tasks. For all methods, the reported model performance in each experiment is measured after $2^{27}$ query molecules had been seen in training, or until the average Pearson's $r$ across all test assays stopped improving on PubChemBA with 8 context compounds. A grid search was performed for all sets of hyperparameters, which are listed below for each model (with the best hyperparameters bolded, according to the highest Pearson's $r$). All models were trained on a server with 8 NVIDIA GTX 3080 GPUs.

\subsection{FS-CAP}
\ours{} was implemented in PyTorch. We used an Adam optimizer with a base learning rate of $10^{-5}$ and 128 steps for learning rate warmup, and then cosine annealed the learning rate to 0 over all training steps. We used dropout ($p=0.1$) and batch normalization following each layer in the predictor network (except in the last 2 layers), while the encoder networks used neither.

Hyperparameters: \texttt{learning\_rate=\{1e-4, 5e-4, 1e-5, \textbf{5e-5}, 1e-6\, 5e-6\}, batch\_size=\{512, \textbf{1024}\}, encoding\_dim=\{256, \textbf{512}\}, n\_layers=\{4, 5, \textbf{6}\}, mlp\_width=\{\textbf{2048}\}}. We used the same number of layers, \texttt{n\_layers}, and width of layers, \texttt{mlp\_width}, in both the context encoder, query encoder, and predictor networks.

\subsection{Tanimoto similarity}
We used 2048-bit Morgan fingerprints with a radius of 3 for the calculation of Tanimoto similarity. When using multiple context compounds, we calculate the Tanimoto similarity between each context compound and all query compounds, but only use the highest similarity context compound for each query compound. This is because if a query compound is similar to one of, but not all, the known actives (the context set), it is still presumed to be active.

\subsection{MolBERT + ANP}
We used the pretrained MolBERT model available from \url{https://github.com/BenevolentAI/MolBERT} to encode SMILES strings into a 768-dimensional vector. We then used this featurizer (which was not made trainable) in an attentive neural process architecture to represent the context and query features, $x_i$ and $x_*$, respectively \cite{kim2019attentive}. We re-implemented the attentive neural process architecture in PyTorch, following the original paper \cite{kim2019attentive} and their published code (\url{https://github.com/deepmind/neural-processes/blob/master/attentive_neural_process.ipynb}) as closely as possible. We trained the model using an Adam optimizer with a base learning rate of $10^{-5}$ and 128 steps for learning rate warmup, and then cosine annealed the learning rate to 0 over all training steps.

Hyperparameters: \texttt{learning\_rate=\{1e-4, \textbf{1e-5}, 1e-6\}, batch\_size=\{\textbf{512}, 1024, 2048\}, num\_attention\_heads=\{2, \textbf{4}, 8\}, encoding\_dim=\{256, \textbf{512}\}, decoder\_layers=\{4, \textbf{5}, 6\}, mlp\_width=\{\textbf{2048}\}}. We use the same \texttt{encoding\_dim} for both the determinstic and latent paths.

\subsection{NGGP}
We used the official PyTorch implementation of NGGP available at \url{https://github.com/gmum/non-gaussian-gaussian-processes}. Using the existing code available for the \texttt{QMUL} dataset, we modified the datalaoders for our task by outputting 2048-bit Morgan fingerprints. We trained only two separate models, one for BindingDB and one for PubChemBA, because the size of the context set is only relevant at test time. We also expanded the \texttt{MLP2} model used in the code to more layers, so the number of parameters was about equivalent to other baselines.

Hyperparameters: \texttt{all\_lr=\{1e-2, \textbf{1e-3}, 1e-4\}, meta\_batch\_size=\{\textbf{5}, 10\}, update\_batch\_size=\{\textbf{5}, 10\}, noise=\{gaussian, \textbf{none}\}, cnf\_dims=\{32, 64, \textbf{128}\}, mlp\_layers=\{4, 5, \textbf{6}\}, nonlinearity=\{\textbf{tanh}, relu\}, batch\_norm=\{True, \textbf{False}\}, mlp\_width=\{\textbf{2048}\}}. We used the defaults provided in the code for the \texttt{QMUL} dataset for all other hyperparameters.

\subsection{MetaNet}
We adapted the Chainer code provided in the official MetaNet implementation (\url{https://bitbucket.org/tsendeemts/metanet/src/master/}) to PyTorch. Most of the architectural choices were kept the same as the original code, although we changed each Block network to include two 2048-wide linear layers with ReLU nonlinearities so that the entire model used about the same number of parameters as other baselines. We trained the model using an Adam optimizer with a base learning rate of $10^{-5}$ and 128 steps for learning rate warmup, and then cosine annealed the learning rate to 0 over all training steps.

Hyperparameters: \texttt{learning\_rate=\{1e-2, \textbf{1e-3}, 1e-4\}, num\_blocks=\{4, \textbf{5}, 6\}, mlp\_width=\{\textbf{2048}\}, hidden\_dim=\{512, 1024, \textbf{2048}\}, batch\_size=\{\textbf{8}, 16, 32\}}

\subsection{MAML}
We used the MAML implementation in the learn2learn library \cite{arnold2020learn2learn}. The base model was a simple multilayer perceptron that takes a 2048-bit Morgan fingerprint as input and produces a single scalar output, which is the activity value prediction. As in the original MAML paper \cite{finn2017maml}, we used an SGD optimizer with a constant learning rate, as well as applied dropout with $p=0.1$ after all layers of the network during training.

Hyperparameters: \texttt{learning\_rate=\{\textbf{1e-4}, 1e-5, 1e-6\}, maml\_learning\_rate=\{\textbf{1e-1}, 1e-2, 1e-3\}, batch\_size=\{512, \textbf{1024}, 2048\}, n\_layers=\{6, 7, \textbf{8}\}, mlp\_width=\{\textbf{2048}\}}

\subsection{MetaDTA}
Since there was no available implementation of MetaDTA, we re-implemented it in PyTorch. For information on the specifics of the MetaDTA architecture, see Section 3.2 of \citet{lee2022metadta}. The context and query inputs, $\mathbf{x}_i$ and $\mathbf{x}_q$ as described in the paper, were represented with 2048-bit Morgan fingerprints, and the context target $y_i$ used the same scalar activity representation as \ours{}. As it does not specify in the original paper, similarly to \ours{}, we used an Adam optimizer with a base learning rate of $10^{-5}$ and 128 steps for learning rate warmup, and then cosine annealed the learning rate to 0 over all training steps.

Hyerparameters: \texttt{learning\_rate=\{1e-4, \textbf{1e-5}, 1e-6\}, batch\_size=\{512, \textbf{1024}, 2048\}, encoding\_dim=\{256, \textbf{512}\}, n\_layers=\{4, 5, \textbf{6}\}, mlp\_width=\{\textbf{2048}\}, attention\_heads=\{1, 2, \textbf{4}, 8\}}. We used the same number of layers, \texttt{n\_layers}, for the query and context set embedding networks, and the decoder network. We also used the same number of attention heads, \texttt{attention\_heads}, for the multi-head cross and self-attention components of the model.

\section{Additional results}
\label{appendix-results}

\subsection{FS-Mol}

\begin{table*}[h]
\caption{\textbf{Regression results on FS-Mol.} We report the mean $\pm$ standard deviation $R^2_{os}$ value across all FS-Mol test tasks, following \citet{chen2022meta}.}
\label{appendix-fsmol}
\begin{center}
\begin{small}
\begin{sc}
\begin{tabular}{l|ccccc}
\toprule
\# context compounds & 16 & 32 & 64 & 128 & 256\\
\midrule
FS-CAP & $0.258 \pm 0.022$ & $0.277 \pm 0.021$ & $0.255 \pm 0.030$ & $0.289 \pm 0.026$ & $0.305 \pm 0.028$\\
\bottomrule
\end{tabular}
\end{sc}
\end{small}
\end{center}
\vskip -0.1in
\end{table*}

Table \ref{appendix-fsmol} reports results on the few-shot regression of activity values from the FS-Mol dataset \cite{stanley2021fs}. While the original FS-Mol paper does not evaluate methods on the regression task, we use the same experimental setting as \citet{chen2022meta}, which is to measure the average task-level out-of-sample coefficient of determination ($R^2_{os}$) across 10 random support/query sets. See \citet{chen2022meta} for comparison with other methods (they provide performance values in a bar chart, so we could not obtain the numeric values for this table).

\subsection{PubChemBA with no activity constraint}

\begin{table*}[h]
\caption{\textbf{Average per-assay correlation.} Mean Pearson's $r$ between predicted and ground-truth compound activity values across all test-set assays in PubChemBA and BindingDB. Context compounds are drawn at random without respect to their activity value. Experiments were performed using 1, 2, 4, and 8 context compounds for each method tested.}
\label{appendix-corr-table}
\vskip 0.15in
\begin{center}
\begin{small}
\begin{sc}
\begin{tabular}{l|cccc|cccc}
\toprule
Dataset & \multicolumn{4}{c}{PubChemBA} & \multicolumn{4}{c}{BindingDB}\\
\# context compounds & 1 & 2 & 4 & 8 & 1 & 2 & 4 & 8\\
\midrule
Tanimoto similarity & 0.00 & 0.11 & 0.20 & 0.29 & -0.04 & 0.06 & 0.15 & 0.18\\
MolBERT + ANP & 0.09 & 0.10 & 0.21 & 0.17 & 0.04 & 0.04 & 0.06 & 0.09\\
NGGP & 0.05 & 0.11 & 0.24 & 0.37 & 0.04 & 0.06 & 0.10 & 0.15\\
MetaNet & 0.01 & 0.01 & -0.01 & 0.00 & 0.02 & 0.05 & -0.01 & 0.01\\
MAML & 0.40 & 0.38 & 0.37 & 0.38 & 0.37 & 0.35 & 0.35 & 0.35\\
MetaDTA & 0.47 & 0.47 & 0.49 & 0.51 & 0.43 & 0.43 & 0.44 & 0.43\\
\midrule
FS-CAP & \textbf{0.51} & \textbf{0.52} & \textbf{0.54} & \textbf{0.54} & \textbf{0.46} & \textbf{0.48} & \textbf{0.49} & \textbf{0.48}\\
\bottomrule
\end{tabular}
\end{sc}
\end{small}
\end{center}
\vskip -0.1in
\end{table*}

Table \ref{appendix-corr-table} reports the same experimental setting as is reported in Section \ref{sec-scoring}, except without any constraints placed on the activity of context compounds. Here, we simply drew context compounds at random, regardless of their activity value. The models trained on this task using the PubChemBA dataset were applied to the tasks presented in Sections \ref{sec-hts} and \ref{sec-cancer}, as the tasks presented in those sections similarly do not have activity constraints on the context compounds.


\end{document}